\documentclass[sigconf]{acmart}

\AtBeginDocument{%
  \providecommand\BibTeX{{%
    \normalfont B\kern-0.5em{\scshape i\kern-0.25em b}\kern-0.8em\TeX}}}

\usepackage[T1]{fontenc}
\usepackage[utf8]{inputenc}

\usepackage{verbatim}
\usepackage{amsmath}
\usepackage{booktabs} 
\usepackage{algorithm}
\usepackage{algorithmicx} 
\usepackage{algcompatible}
\usepackage{algpseudocode}
\usepackage{multirow}
\usepackage{color}
\usepackage{threeparttable}
\usepackage{pifont}
\usepackage{etaremune}
\usepackage{caption}
\usepackage{subcaption}
\usepackage[export]{adjustbox}
\usepackage{tablefootnote}
\usepackage{makecell}

\usepackage{url}
\usepackage{enumitem}
\usepackage{todonotes}

\def\BibTeX{{\rm B\kern-.05em{\sc i\kern-.025em b}\kern-.08emT\kern-.1667em\lower.7ex\hbox{E}\kern-.125emX}}

\newcommand{\ten}[1]{\mathcal{#1}}
\newcommand{\mat}[1]{\mathbf{#1}}

\copyrightyear{2021} 
\acmYear{2021} 
\setcopyright{acmcopyright}
\acmConference[GLSVLSI '21]{Proceedings of the Great Lakes Symposium on VLSI 2021}{June 22--25, 2021}{Virtual Event, USA}
\acmBooktitle{Proceedings of the Great Lakes Symposium on VLSI 2021 (GLSVLSI '21), June 22--25, 2021, Virtual Event, USA}
\acmPrice{15.00}
\acmDOI{10.1145/3453688.3461738}
\acmISBN{978-1-4503-8393-6/21/06}
\begin{document}

\title{
3U-EdgeAI: Ultra-Low Memory Training, Ultra-Low Bitwidth Quantization, and Ultra-Low Latency Acceleration
}

\author[]{Yao Chen }
\affiliation{%
\small
 yao.chen@adsc-create.edu.sg\\
  \institution{Advanced Digital Sciences Centre, Singapore}
  \country{}
}
\author[]{Cole Hawkins}
\affiliation{%
\small
colehawkins@ucsb.edu \\
  \institution{University of California, Santa Barbara, USA}
  \country{}
}
\author[]{Kaiqi Zhang}
\affiliation{%
\small
kzhang70@ucsb.edu \\
  \institution{University of California, Santa Barbara, USA}
  \country{}
}
\author[]{Zheng Zhang}
\affiliation{%
\small
zhengzhang@ece.ucsb.edu \\
  \institution{University of California, Santa Barbara, USA}
  \country{}
}
\author[]{Cong Hao}
\affiliation{%
\small
callie.hao@ece.gatech.edu \\
  \institution{Georgia Institute of Technology}
  \country{}
}




\begin{abstract}
The deep neural network (DNN) based AI applications on the edge require both low-cost computing platforms and high-quality services.
However, the limited memory, computing resources, and power budget of the edge devices constrain the effectiveness of the DNN algorithms.
Developing edge-oriented AI algorithms and implementations (e.g., accelerators) is challenging.
In this paper, we summarize our recent efforts for efficient on-device AI development from three aspects, including both training and inference.
First, we present on-device training with \textbf{ultra-low memory} usage. We propose a novel rank-adaptive tensor-based tensorized
neural network model, which offers orders-of-magnitude memory reduction during training.
Second, we introduce an \textbf{ultra-low bitwidth} quantization method for DNN model compression, achieving the state-of-the-art accuracy under the same compression ratio.
Third, we introduce an \textbf{ultra-low latency} DNN accelerator design, practicing the software/hardware co-design methodology.
This paper emphasizes the importance and efficacy of training, quantization and accelerator design, and calls for more research breakthroughs in the area for AI on the edge.
\end{abstract}



\maketitle

\section{Introduction}

Deep neural networks (DNNs) are becoming attractive solutions for many edge AI applications and have made remarkable progress in various areas such as computer vision, natural language processing, health care, autonomous driving, and surveillance.
Meanwhile, with the increase of the size and complexity of the neural networks, training and deploying a DNN with a large number of parameters and complex data transmission on small and power-constrained edge devices, such as smart phones and wearable devices, becomes increasingly challenging~\cite{hao2021enabling, han2015deep,zhang2017machine}.
In this work, we focus on three primary challenges: \textbf{ultra-low memory} training, \textbf{ultra-low bitwidth} quantization, and \textbf{ultra-low latency} acceleration, and discuss our solutions for each of them.

First, there is an increasing demand for on-device machine learning model training, to preserve data privacy, enable model personalization and lifelong learning, and to improve energy efficiency to avoid the massive data transmission to the cloud~\cite{7979979, wang2019e2}. 
However, model training has a much larger memory requirement than inference, exposing additional challenges for on-device training, where the edge-devices are usually equipped with limited memory capacity.
Therefore, \textbf{ultra-low memory} training method must be explored to enable on-device training.
To this end, we present \textit{an end-to-end low-precision tensorized neural network training framework} with orders-of-magnitude memory reduction~\cite{zhang2021fpga}. The rank-adaptive tensorized training method employs a Bayesian method for automatic tensor rank determination and model compression in the training process.

Second, to implement DNNs on the memory-constrained edge devices,
pruning and quantization are promising to reduce the number of weights and the data bit-width in DNN models, with an extreme case that quantizes the weights down to binary/ternary representations~\cite{han2015deep, li2016ternary, Courbariaux-binary}.
These methods can dramatically reduce the network size as well as number of the multiplications during the execution of the model. 
Given the tight memory and computing resource budget on the edge, \textbf{ultra-low bitwidth} quantization methods are especially attractive.
However, ultra-low bitwidth quantization can easily cause significant degradation on the model accuracy, making such aggressive quantization methods challenging.
To address such challenges, we present \textit{a novel ternary weight quantization method by proposing a vectorized loss function}, achieving the state-of-the-art accuracy under the same compression ratio~\cite{gong2020vecq}.

Third, for efficient DNN deployment on the edge-devices, FPGAs are becoming attractive platforms comparing with CPUs, GPUs and digital signal processors (DSPs)~\cite{obtract,zhang2018dnnbuilder,qiu2016going}.
FPGAs can provide the flexibility to be configured as domain specific architecture that can meet various implementation requirements such as \textbf{ultra-low latency} on the edge-devices.
In addition, modern SoC FPGAs integrate low power processors and sufficient interfaces that can support widely used sensors for Internet-of-things (IoT) applications.
We present \textit{the first instruction based ternarized low-latency deep learning accelerator} with high performance, low resource utilization, and high flexibility for different DNN models~\cite{chen2019tdla}.



The remaining of this paper is organized as follows.
Section~\ref{sec:low-memory} introduces our low-memory rank-adaptive on-device training framework;
Section~\ref{sec:low-bitwidth} introduces our low-bitwidth DNN quantization solution;
Section~\ref{sec:low-latency} introduces our low-latency DNN accelerator design.
In Section~\ref{sec:experiments} we demonstrate the effectiveness of our proposed methods, followed by the conclusions and future work in Section~\ref{sec:conclusion}. 
\section{Ultra-Low Memory Training}
\label{sec:low-memory}

The large amount of model parameters consume massive computing and memory resources, which prevents direct training of neural networks on edge devices. A promising technique of reducing model parameters is low-rank tensor decomposition~\cite{kolda2009tensor,oseledets2011tensor}.  This method has achieved great success in post-training compression and fixed-rank training~\cite{zhou2019tensor, calvi2019tucker, yin2020compressing, tjandra2017compressing,lebedev2014speeding,novikov2015tensorizing,garipov2016ultimate}. However, several fundamental issues need to be addressed in on-device one-shot training:
\begin{itemize}[leftmargin=*]
    \item Firstly, a rank-adaptive training framework is needed to avoid combinatorial search of tensor ranks and multiple training runs. 
    \item Secondly, hardware-friendly tensor algorithms should be developed to facilitate their implementation on edge devices. 
\end{itemize}
In this section, we summarize our recent work on the algorithm~\cite{hawkins2019bayesian,hawkins2020towards} and hardware~\cite{zhang2021fpga} levels to address these challenges.

\begin{figure}[t]
\centering
  \begin{subfigure}[t]{0.22\textwidth}
  \centering
    \includegraphics[width=0.35\textwidth]{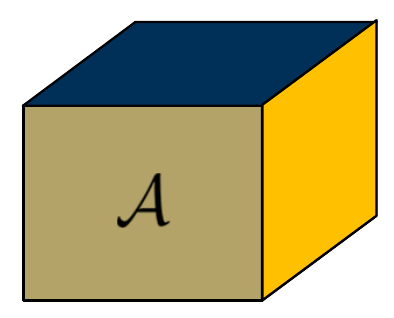}
    \caption{\label{fig: full visual}}
  \end{subfigure}
\begin{subfigure}[t]{0.25\textwidth}
  \centering
    \includegraphics[width=1.0\textwidth]{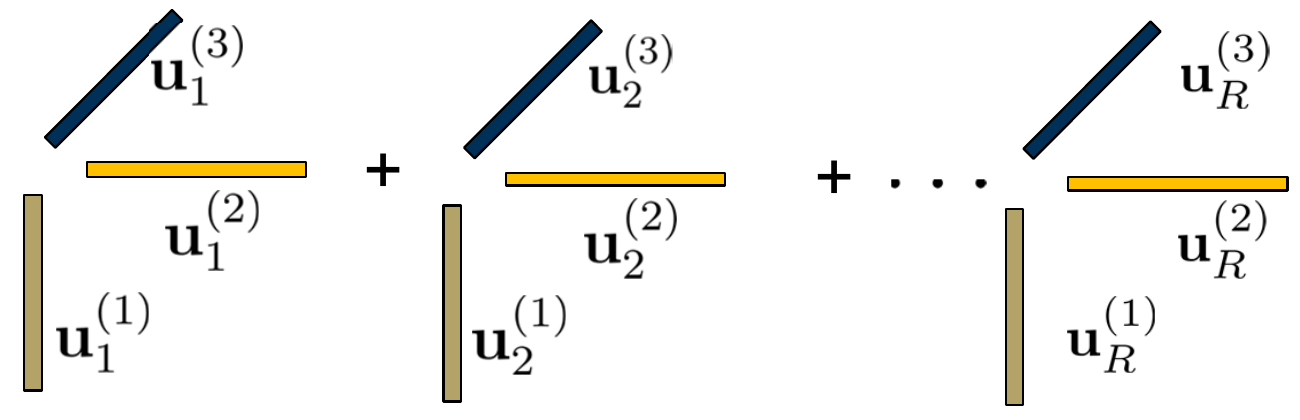}
    \caption{\label{fig: cp visual description}}
  \end{subfigure}
  \\

     \begin{subfigure}[t]{0.22\textwidth}
  \centering
    \includegraphics[width=.65\textwidth]{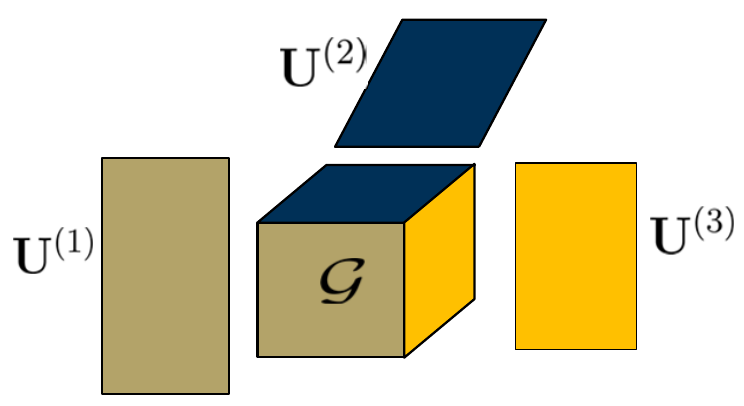}
    \caption{\label{fig: tucker visual description}}
  \end{subfigure}
     \begin{subfigure}[t]{0.25\textwidth}
  \centering
    \includegraphics[width=1\textwidth]{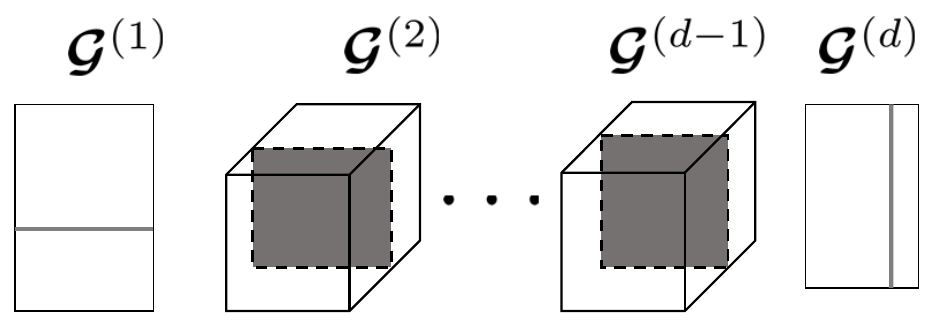}
    \caption{\label{fig: tt visual description}}
  \end{subfigure}
  \vspace{-6pt}
  \caption{(a): An order-3 tensor. (b) and (c): CP and Tucker representations, respectively. (d): TT representation, where the gray lines and squares indicate a slice of the TT core by fixing its mode index. This figure is reproduced from~\cite{hawkins2020towards}. \label{fig: visual formats}}
  \vspace{-8pt}
\end{figure}

\subsection{Bayesian Tensorized Training Models}

\subsubsection{ Low-rank tensor representation.} In many cases we can describe a neural network with much less parameters via low-rank tensors. Consider a weight matrix $\mat{W}\in\mathbb{R}^{J\times I}$ for example (and other parameters such as convolutional filters and embedding tables can be handled similarly). We can firstly fold $\mat{W}$ to a high-dimensional tensor $\ten{W}$ of size $J_1\times\dots\times J_d\times I_1\times\dots\times I_d$, where $I=\prod_{n=1}^d I_n,J=\prod_{n=1}^d J_n$. Then, we can describe the tensor $\ten{W}$ with some low-rank tensor factors $\boldsymbol{\Phi}$. This can be done with various low-rank tensor decomposition formats as shown in Fig.~\ref{fig: visual formats}~\cite{hawkins2020towards}. In various tensor decompositions, $\boldsymbol{\Phi}$ denotes the associated tensor factors. For large fully connected layers and embedding tables, the tensor-train matrix (TTM) format turns to be highly effective~\cite{hawkins2020towards}. In the TTM format, $\boldsymbol{\Phi}= \{\ten{G}^{(n)}\}_{n=1}^d$, and each $\ten{G}^{(n)}\in \mathbb{R}^{R_{n-1}\times J_n \times I_n \times R_{n} }$ is an order-$4$ TTM core. The vector $\mat{R}=(R_0,R_1,\cdots,R_d)$ with $R_0=R_d=1$ is the tensor ranks that determine the model complexity. With low-rank tensors, one may reduce the number of model parameters from an exponential function of $d$ to a linear one. 

\subsubsection{Bayesian Tensorized End-to-End Training.} Despite the high compression ratio via tensor methods, determining the tensor rank in advance is very hard~\cite{hillar2013most}. This is further complicated by the nonlinear forward model in neural networks, which has prevented tensorized one-shot on-device training in previous works. We have developed two Bayesian models to address this issue:
\begin{itemize}[leftmargin=*]
    \item {\bf Stein Variational Inference for TTM Format.} In~\cite{hawkins2019bayesian}, we have considered TTM format. We model each slice of $\ten{G}^{(k)}$ with a zero-mean Gaussian prior density. We further control the variance by two tunable Gamma hyper-priors to enforce low tensor ranks. The actual tensor rank is decided jointly by the training data and rank-controlling hyper-parameters. Starting from an initial rank parameter $R_k$, we can learn an actual rank $\hat{R}_k \leq R_k$, leading to further model compression in the training process. This method uses a Stein variational inference~\cite{liu2016stein} to compute the posteior density for small- or medium-size neural networks. 
    
    \item {\bf Scalable SVI for One-Shot Tensorized Training. } In~\cite{hawkins2020towards}, we have developed a more generic and efficient Bayesian model for tensorized training. This work can handle CP, Tucker, TT and TTM formats. It uses Gaussian priors to model low-rank tensor factors, and uses Half-Cauchy or Log-Uniform hyper-priors to control tensor ranks. We have improved the stochastic variational inference (SVI) ~\cite{hoffman2013stochastic} by two steps. Firstly, we simplify the posterior density of rank-controlling hyper-parameters to a Delta function to avoid gradient explosion. Secondly, we use a hybrid numerical/analytical update rule inside SVI. This highly scalable method can perform one-shot training of very large-scale neural networks with billions of model parameters. 
\end{itemize}

\subsubsection{Performance Summary.} 
\begin{itemize}[leftmargin=*]
    \item Our first method~\cite{hawkins2019bayesian} has been tested on a two-layer fully connected neural network, a 6-layer CNN and a 110-layer residual neural network. Our work has produced $7.4\times $ to $137\times $ more compact neural networks directly from the training with little or no accuracy loss.
    
    \item Our recent work~\cite{hawkins2020towards} has been tested on a practical CNN, a large-scale NLP model~\cite{khrulkov2019tensorized} and an extremely large deep learning recommendation model (DLRM)~\cite{naumov2019deep} from Facebook. Orders-of-magnitude parameter reduction has been achieved in the training process. As shown in Table~\ref{tab:dlrm}, training the DLRM with a standard method involves $4.25\times 10^9$ variables. Our proposed method only trains $2.36 \times 10^6$ variables due to low-rank tensorization, and it further reduce the model parameters to $164$K in the training process due to the automatic rank determination. The overall parameter reduction ratio in the training process is $2.6\times 10^4$.
\end{itemize}

\begin{table}[t]
\caption{Performance of our tensorized training~\cite{hawkins2020towards} on the Facebook DLRM model.}
\small
\begin{tabular}{|c|c|c|c|}
\hline
              & standard & tensorization & rank-adaptive training \\ \hline
\# parameters & 4.25B    & 2.36M         & 164K                   \\ \hline
compression   & N/A      & $1,800\times$          & $26,000\times$       \\ \hline           
\end{tabular} \normalsize
\label{tab:dlrm}
\vspace{-10pt}
\end{table}


\subsection{One-Shot On-Device Tensorized Training}
To demonstrate on-device training, we have developed a low-precision tensorized training algorithm and its FPGA prototype~\cite{zhang2021fpga}.

\subsubsection{Low-Precision Tensorized Training.} We consider the maximum a posteriori probability (MAP) estimate of the Bayesian model \cite{hawkins2020towards}. In this case, the training loss function includes two parts:
the cross-entropy loss of a neural network classifier dependent on TTM factors $\{ \ten{G}^{(k)}\}_{k=1}^d$, and a regularization term caused by the Gaussian priors of TTM factors as well as the Log-Uniform hyper-priors for rank-controlling parameters $\boldsymbol{\lambda}_k$'s. In the training process, both TTM factors and rank-controlling parameters will be computed. To reduce the training cost on hardware, a low-precision tensorized training algorithm is developed based on the following key ideas:
\begin{itemize} [leftmargin=*]
    \item We use BinaryConnect~\citep{courbariaux2015binaryconnect} to compute low-precision TTM factors. BinaryConnect keeps the real values of all low-precision parameters in a buffer.  In each iteration, the gradients are accumulated in the buffer, and the low-precision parameters are updated by quantizing the buffer. To handle the non-differentiable quantization function in the training process, we use the straight-through estimator (STE) \citep{bengio2013estimating} to approximate its gradient. 
    \item We use different precisions for different variables in the training process. Specifically, we use 4 bits to represent TT factors, 8 bits for activations and bias, and 16 bits for the gradients.
\end{itemize}

\subsubsection{On-FPGA Training.} 
\begin{figure}[t]
	\centering
		\includegraphics[width=0.45\textwidth]{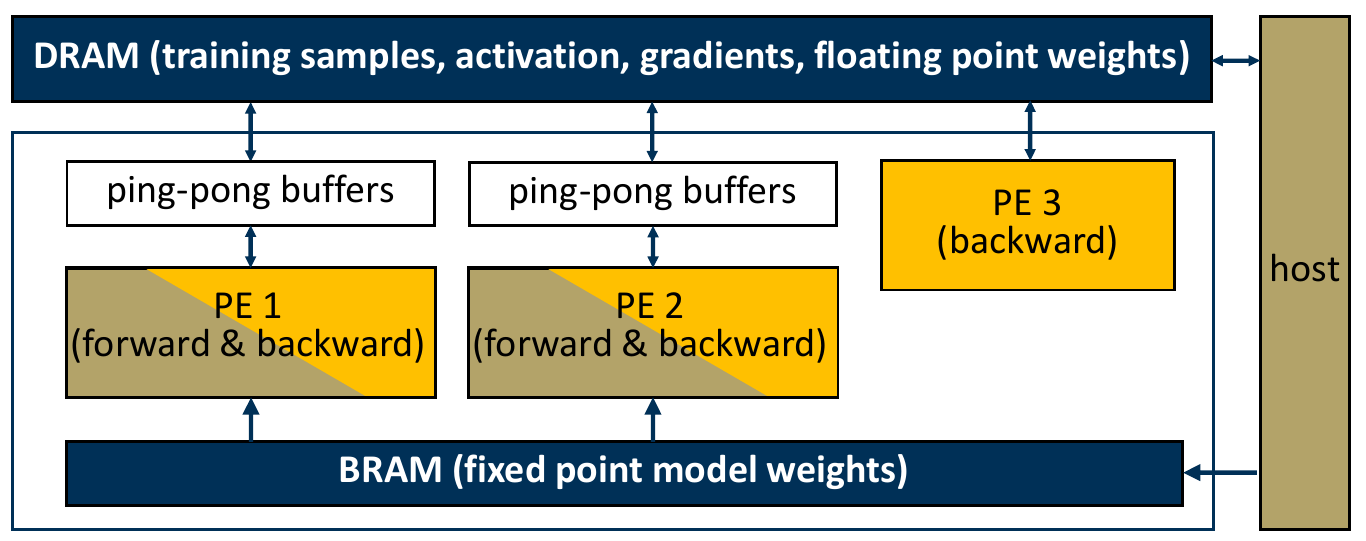} 
		\vspace{-6pt}
\caption{Our FPGA accelerator for the end-to-end tensorized training. Reproduced from~\cite{zhang2021fpga}.}
\vspace{-10pt}
\label{fig:fpga}
\end{figure}

To demonstrate our training algorithms on edge devices, we have implemented an FPGA accelerator as shown in Fig.~\ref{fig:fpga} for the low-precision tensorized training framework. 
\begin{itemize}[leftmargin=*]
    \item Since our low-rank tensorization can greatly reduce the training variables, all model parameters may be stored in the on-chip BRAM. The data samples, activations, and gradients are stored in the off-chip DRAM during the training process.
    \item The forward and backward propagations are run on the FPGA programmable logic.
The TTM factors and rank-controlling parameters are updated on the embedded ARM core. 
\item Three processing elements (PEs) are designed for the forward and backward propagation. 
PE1 and PE2 are shared by the forward and backward propagations, and they handle tensor contractions. PE1 is used for a two-index tensor contraction which contains the last dimension of two tensor variables. In contrast, PE2 performs a tensor contraction along a single dimension that is not the last. PE3 computes the outer products in a backward propagation. 
\end{itemize} 

\section{Ultra-low Bitwidth Quantization}
\label{sec:low-bitwidth}

Neural network quantization employs low precision (bitwidth) data for efficient model execution.
Especially, ultra-low bitwidth quantization leads to much less memory usage, lower complexity of the multiply-accumulate operations, and higher efficiency of model execution, making it an appealing technology for enabling AI at edge devices.
However, aggressively lowering the data bitwidth (e.g., lower than 4-bit) is very challenging:
\begin{itemize}[leftmargin=*]
    \item {It can easily result in large accuracy degradation~\cite{cong2019dac,chen2019tdla,gysel2016hardware}, requiring a careful balance between the computing efficiency and the final model accuracy.
    }
    \item {
    Minimizing the quantization loss, i.e., the L2 distance between the original and the quantized values, is an appealing method~\cite{han2015deep,ENN2017,TSQ2018,cheng2019uL2Q, li2016ternary, leng2018extremely} but have major drawbacks such as easily falling into local optima and neglecting the distribution and correlations of the weights~\cite{gong2020vecq}.
    }
\end{itemize}

To address such challenges and achieve high-accuracy ultra-low bitwidth quantization, we have proposed a quantization method, namely \textbf{VecQ}~\cite{gong2020vecq}, with a novel \textit{vectorized loss function} and an open-sourced training flow.
VecQ can quantize the model weights into 1-bit to 16-bit and shows exceptional performance especially under ultra-low bitwidth, e.g., ternary values.

\begin{figure}[t]
    \centering
    \includegraphics[width=0.22\textwidth]{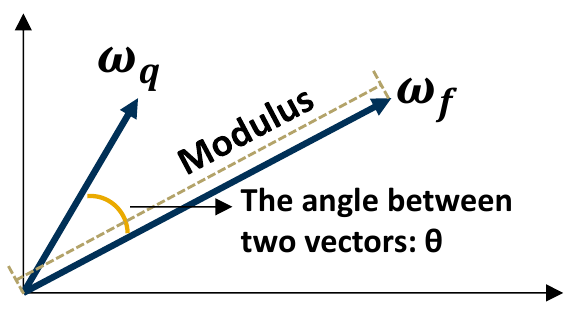}
    \vspace{-9pt}
    \caption{The quantization angle between $\omega_f$ and $\omega_q$.}
    \vspace{-9pt}
    \label{fig:angle}
\end{figure}

\textbf{Vectorized Loss Function.}
We organize the weights within one DNN layer into a vector $\omega \in \mathbb{R}^{l}$ where $l$ is the number of weights,
and denote the original floating point weight vector by $\omega_f$ and the quantized weight vector by $\omega_q$.
Typically, there is a scaling factor $\alpha$ such that $\omega_q = \alpha \boldsymbol{v}$, where each element in $\boldsymbol{v}$ is a low-bitwidth representation.
Based on the vector representations, we define the \textit{quantization angle} between $\omega_f$ and $\omega_q$, denoted by $\theta$.
Figure~\ref{fig:angle} shows an example when $l=2$.
The objective is to find optimal $\alpha$ and $\boldsymbol{v}$ such that $\omega_q$ is as close to $\omega_f$ as possible.

We propose the \textit{vectorized loss} to describe the quantization loss, denoted by $J_v$, and we minimize $J_v$ during the quantization.
$J_v$ is defined as the summation of the \textit{orientation loss}, denoted by $J_o$, and the \textit{modulus loss}, denoted by $J_m$ as follows:
\begin{equation}
    \begin{split}
    J_o&=1-\cos\theta, \quad \text{where} \;\cos\theta=\frac{\alpha \boldsymbol{v} \cdot \omega_f } {|| \alpha \boldsymbol{v} || \cdot || \omega_f||} \\
    J_m&=|| \omega_f- \alpha \boldsymbol{v} ||_2^2 \\
    J_{v} &= J_o + J_m
\end{split}
\end{equation}
The orientation loss describes the angle between two vectors, while the modulus loss describes the squared distance between $\omega_f$ and $\omega_q$. 
Notably, by minimizing $J_v$, we usually achieve lower quantization loss comparing with directly minimizing $J_m$. More details can be found in the VecQ paper~\cite{gong2020vecq}.

\begin{figure}
     \centering
     \includegraphics[width=0.37\textwidth]{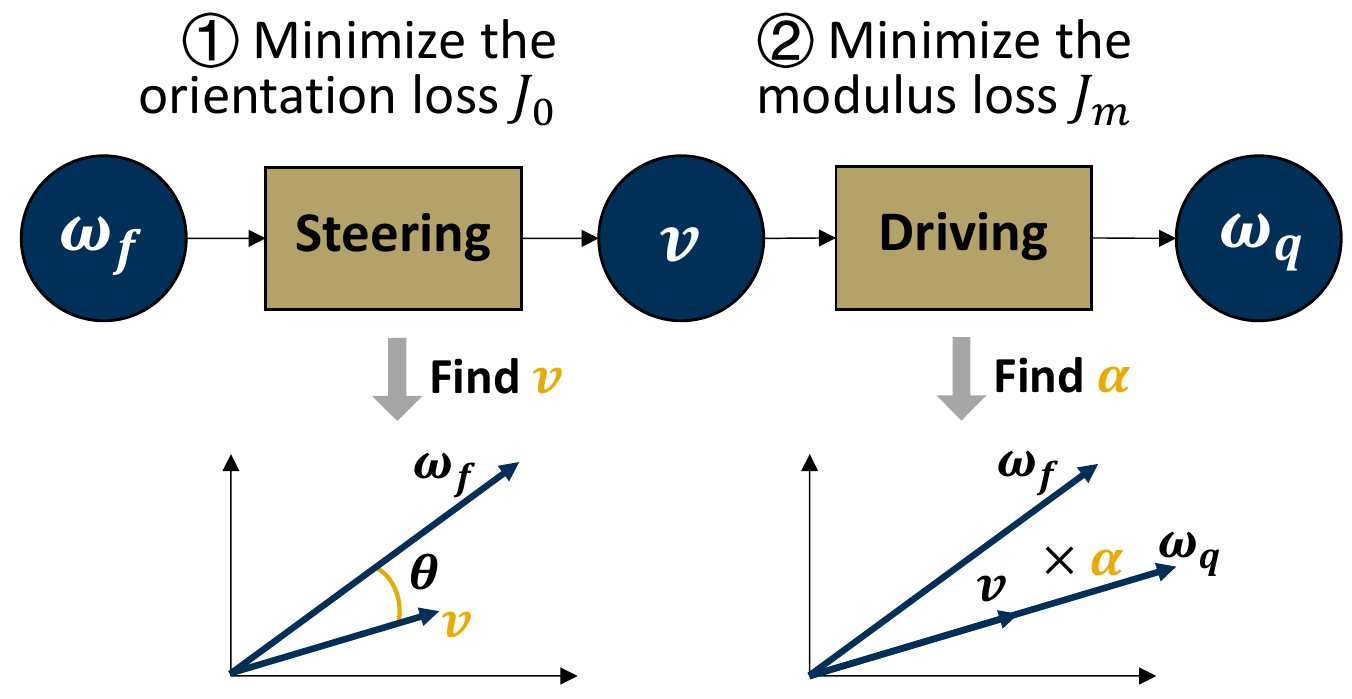}
     \vspace{-6pt}
     \caption{The data quantization flow of VecQ, reproduced from VecQ~\cite{gong2020vecq}.}
     \label{fig:vec}
\end{figure}

\textbf{Vectorized Loss Minimization.}
We minimize $J_v$ in two steps, namely steering and driving, as shown in Figure~\ref{fig:vec}.
First, the steering step minimizes the orientation loss $J_o$ to find the best $\boldsymbol{v}$, since $J_o$ is independent of $\alpha$.
Second, the driving step minimizes the modulus loss $J_m$ to find the best scaling factor $\alpha$. 
For a convolution layer, all the weights within the same layer share the same scaling factor; for a depth-wise convolution layer, each kernel has its own scaling factor for a better representation of the less number of weights for it.
We quantize the activations to fixed point values during the training to further reduce the memory utilization.

\textbf{Training Flow Integration.}
We integrate our VecQ solution into the Tensorflow and Pytorch DNN training frameworks.
For each layer, in the forward propagation, 
we first quantize the weights from $\omega_f$ to $\omega_q$, and then use $\omega_q$ to compute the output activations, which is also quantized into fixed point.
In the backward propagation, the gradients are also calculated using $\omega_q$ to update $\omega_f$.


\section{Ultra-low Latency Acceleration}
\label{sec:low-latency}

The effectiveness of a dedicated FPGA accelerator for DNN models have been widely demonstrated~\cite{qiu2016going, chen2019clouddnn}. 
However, ultra-low latency accelerators for edge devices with an extremely limited resource budget still require careful design considerations.

Benefiting from our quantization solution VecQ for ultra-low bitwidth,
we have proposed \textbf{T-DLA}, a light-weight ternarized accelerator overlay under strict resource constraint, to achieve ultra-low latency on edge devices~\cite{chen2019tdla}.
The key features of T-DLA include:
\begin{itemize}[leftmargin=*]
    \item An optimized and expressive single instruction multiple data (SIMD) instruction set.
    \item A novel memory sub-system supporting effective data access of the computation modules.
    \item An efficient execution pipeline with low-latency computation modules.
\end{itemize}


\begin{table}[t]
\footnotesize
\caption{The format of the instruction word in T-DLA.}\label{tab:inst}
\begin{tabular}{|l|l|l|l|l|l|l|l|l|}
\hline
Byte Idx    & 7  & 6  & 5   & 4   & 3   & 2   & 1  & 0  \\ \hline
Load & OP & FS & SAM & SAL& DAM & DAL  & KS  & CC  \\ \hline
\end{tabular}
\\
\footnotemark[1]{OP: operation code~~~}
\footnotemark[2]{FS: input feature size}\\
\footnotemark[3]{SAM/SAL: source address most/least significant byte}\\
\footnotemark[4]{DAM/DAL: destination address most/least significant byte}\\
\footnotemark[5]{KS: kernel size~~~}
\footnotemark[6]{CC: in/out/activation/pooling selection}
\vspace{-9pt}
\end{table}

\textbf{SIMD Instruction Set.}
To support the task scheduling for various DNN models,
the instruction set of T-DLA is designed as simple yet expressive enough for a large variety of DNNs.
Each instruction is a 64-bit word (8 bytes) with the format shown in Table~\ref{tab:inst}. 
The payloads of the different bytes are generated according to the layer configurations.



\begin{figure}[]
	\centering
	\includegraphics[width=0.46\textwidth]{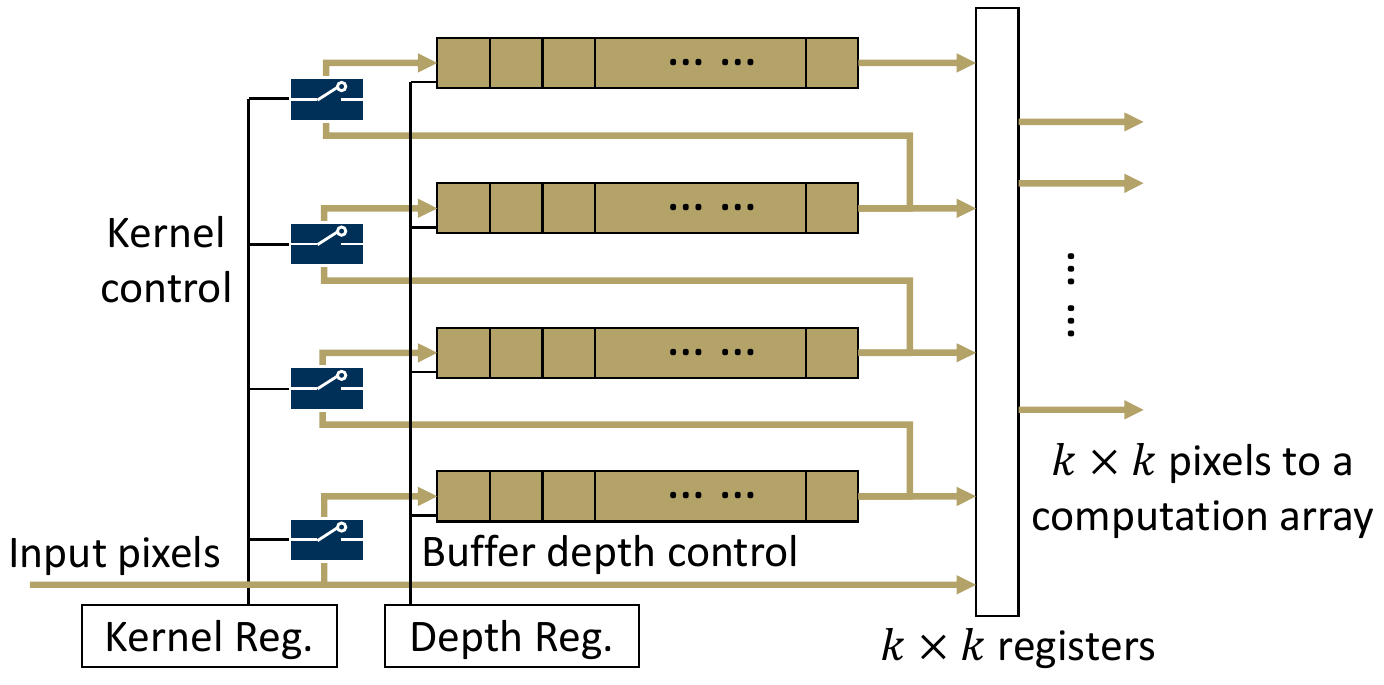}
	\vspace{-3pt}
	\caption{The variable-length line buffer. The kernel size and the depth of the buffers are both configurable.}
	\label{fig:linebuf}
	\vspace{-8pt}
\end{figure}

\textbf{Memory Sub-system.}
The memory subsystem contains two levels of storage to provide low latency data fetching to the computation units: a simple input buffer and a variable-length line buffer. The simple input buffer is a BRAM buffer for temporary input feature storage; the variable line buffer serves for the efficient data streaming into the ternary computation array, as shown in Figure~\ref{fig:linebuf}.
It is designed to support variable kernel size $k$ and variable buffer depth $d$, which are specified by the instruction to reduce the data transmission latency caused by fixed hardware paths. Once configured by the instruction, it provides an output of $k \times k$ data each clock cycle to the computation array.

\textbf{Execution Pipeline and Computation Modules.}
T-DLA has four major computation modules: 1) a ternary computation array, 2) a set of adder trees, 3) activation and scaling modules, and 4) pooling modules.

\textit{Ternary computation array.}
With our ternarized model training via VecQ, the weights are represented by 2 bits using two's complement encoding, so that the multiplication in the convolution layer is simplified to selection and inversion logic.
Benefiting from such simplified logic, we can achieve parallelism along the input channel, the output channel, and the kernel dimensions.
The computation array is constructed by $T_{n} \times T_{m} \times L_{K}^2$ computation units, which can process this number of input data simultaneously. $T_{n}$ and $T_{m}$ are the maximum numbers of the input and output channel that can be processed by the computation array, and $L_{K}$ is the pre-defined maximum allowable kernel size.
The values of $T_{n}, T_{m}, L_{K}$ and the length of the line buffer $L_{D}$ are all configurable and could be determined based on the on-chip resource availability. 

\textit{Adder tree.}
Since the computation array is built only using LUTs and FFs, we use DSPs to construct adder trees.
We take advantage of the SIMD mode of the DSPs where the internal carry propagation between segments is blocked to ensure independent operation. 
Therefore, we split the 48-bit input of a DSP into four 12-bit independent accumulation channels, 
so that a single DSP can perform addition for 8 pieces of input data and provide 4 outputs. 
Benefiting from the SIMD mode, the DSPs can provide outputs in every single clock once the internal register lines are filled up.
Furthermore, the clock frequency of the DSPs are configured to be higher than other logic parts with the help of input/output asynchronous FIFOs, which further reduces the processing latency.

\textit{Other modules.}
The ReLU activation module, the linear scaling module, and the max pooling module are all designed to process in a single clock cycle to reduce the depth of the execution pipeline.

\section{Experimental Results}
\label{sec:experiments}

In this section we demonstrate the effectiveness of our methods, including the ultra-low memory training framework, the ultra-low bitwidth VecQ quantization, and the ultra-low latency T-DLA design.
Notbly, all these works are open-sourced.


\begin{table}[t]
    \centering
    \footnotesize
    \renewcommand{\arraystretch}{0.7}
    \setlength{\tabcolsep}{3pt}
    \caption{The on-device ultra-low memory training method on the Fashion MNIST dataset.}
    \begin{tabular}{c|ccccc}
    \hline
    Method & \thead{Training\\ accuracy} & \thead{Testing\\ accuracy} & \thead{Model \\parameters} & \thead{Memory\\ in bits} & \thead{Memory\\reduction}\\
    \hline
    Vanilla & 95.75\% & 89.27\% & $4.67\times 10^5$ & $1.49\times 10^7$ & N/A\\
    \hline
    Floating, w/o prior& 92.54\% & 88.03\% & $1.48\times 10^4$ & $4.74\times 10^5$ & 31.4$\times$\\
    Fixed, w/o prior & 88.31\% & 86.67\% & $1.48\times 10^4$ & $6.13\times 10^4$ & 243$\times$\\
    \hline
    Floating, w/ prior & 90.17\% & 87.88\% & $1.08\times 10^4$ & $3.46\times 10^5$ & $43.1\times$\\
    \textbf{Fixed, w/ prior} & 85.45\% & 84.86\% & $1.22 \times 10^4$ & $5.11\times 10^4$ & 292$\times$\\
    \hline
    \end{tabular}
    \label{tab:fm}
\end{table}

\subsection{On-device Training}

We implement our low-precision rank-adaptive tensorized training on an Avnet Ultra96-V2 FPGA board and use it to train a two-layer neural network for a classification task on the FashionMNIST dataset. 
There are 512 neurons in the hidden layer, folded into $4\times 4 \times 2 \times 16$ for the first layer, and $32 \times 16$ for the second layer.  We  use  the  Pytorch  and  Tensorly  modules  to implement our training algorithm on the embedded processor.  For the FPGA we set the clock rate to 100MHz. We compare the training methods with or without the low rank TT priors. 
As shown in Table \ref{tab:fm}, our method achieves $294\times$ memory reduction for the model parameters compared with the standard non-tensorized training. This on-FPGA training has achieved $59\times$ speedup and $123\times$ energy reduction than the training on an embedded CPU.

\subsection{Ultra-low Bitwidth Quantization}

\begin{table}[]
\centering
\footnotesize
\caption{VecQ: top-1 classification accuracy}
\label{tab:accuracy}
\begin{tabular}{l|l|l|l|l}
\hline
Dataset & MNIST & CIFAR10&CIFAR10&ImageNet \\ \hline
Model & Lenet-5 & Cifarnet & VGG-like& Resnet-18     \\ \hline
Floating     & 99.41  & 80.54  &93.49 & 69.60  \\ \hline
IJCNN'17~\cite{efficient} & 98.33&-&87.89& -\\ \hline
NIPS'16~\cite{li2016ternary} & 99.35&-&92.56& 61.8\\ \hline
\textbf{Ours} & 99.5  & 78.7  &92.94 & 68.23     \\ \hline
\end{tabular}
\vspace{-10pt}
\end{table}

We use MNIST, Cifar10 and ImageNet to evaluate the ultra-low bitwidth quantization, VecQ.
The evaluated DNN models include Lenet-5, Cifarnet, a VGG-like network~\cite{hperf}, and Resnet-18.

\begin{table}[]
\centering
\footnotesize
\caption{VecQ: model size reduction}
\label{tab:size}
\begin{tabular}{l|l|l|l|l}
\hline
Model & Lenet-5 & Cifarnet & VGG-like& Resnet-18     \\ \hline
Param. Total (M)& 0.43 &0.279 & 5.35 & 11.69 \\ \hline
Param. Conv (M)& 0.025&0.258 &1.114 & 11.177 \\ \hline
Floating (MB)     & 1.644  & 1.065 & 20.408 & 44.594   \\ \hline
\textbf{Ours} (MB)   & 0.393& 0.081 & 4.284 & 3.154     \\ \hline
Mem.Reduc.(\%) & 76.09 & 92.39 & 79.01 & 92.93 \\ \hline
\end{tabular}
\end{table}

\begin{table}[]
	\centering
	\footnotesize
	\caption{T-DLA resource and performance}
	\label{tab:perandres}
	\begin{tabular}{l|l|l|l|l}
		\hline
		\thead{Configuration Parameters \\ $<T_{n}, T_{m}, L_{K}, L_{D}, D_{w}>$} & \multicolumn{4}{c}{$<4,16,5,32,12>$} \\ \hline
		\thead{Resource Utilization(\%) \\ $<LUT/FF/BRAM/DSP>$ }   &  \multicolumn{4}{c}{79 / 47.47 / 68.93 / 91.82} \\ \hline
		\thead{Clock Frequency\\ Logic / Adder (MHz)} &  \multicolumn{4}{c}{125 / 250} \\ \hline
		Peak Performance (GOPS) & \multicolumn{4}{c}{400} \\ \hline
		DNN Model        &  Lenet-5 & Cifarnet & VGG-like & Resnet-18 \\ \hline
		latency(ms) & 0.016 &0.063&2.12&48.8 \\ \hline
	\end{tabular}
\end{table}

\begin{table}[]
	\centering
	\footnotesize
	\caption{T-DLA: Comparison with the state-of-the-art implementations.}
	\label{tab:hocom}
	\begin{tabular}{l|l|l|l|l|l|l}  \hline
		Dataset  & Design & Model & Acc.(\%) & F., W. (bits) & fps    & platform               \\ \hline
		MNIST   & \cite{finn} & MFC-max& 97.69 & 1, 1 & 6238000  &  ZC706 \\ \hline
		MNIST   & \cite{impl16} &Lenet-5& - & 8, 3  & 70000  &  ZC706 \\ \hline
		MNIST   & \textbf{Ours}  & Lenet-5 & \textbf{99.5} &8, 2& 62051.1 & Zedboard \\ \hline
		\hline
		CIFAR10 & \cite{finn} & VGG-like& 80.1 & 24, 1 & 21900& ZC706    \\ \hline		
		CIFAR10 & \cite{fcfree} & VGG-like & 81.8 & 1, 1& 420 & Zedboard \\ \hline
		CIFAR 10 & \cite{hperf} & VGG-like& 86.71 & 8, 2& 27043 & VC709    \\ \hline		
		CIFAR 10 & \cite{accbnn}   & VGG-like     & 88.68  & 1, 1& 168 & Zedboard \\ \hline
		CIFAR 10 & \textbf{Ours} & VGG-like & \textbf{89.08} & 8,2  &457& Zedboard \\ \hline
		\hline
		ImageNet & \cite{li2016ternary}&Resnet-18&65.44&FP32,FP32&1.545& Xeon\footnotemark \\ \hline
		ImageNet & \cite{li2016ternary}&Resnet-18&65.44&FP32,FP32&387.597& 1080Ti\footnotemark \\ \hline
		ImageNet &\textbf{Ours} & Resnet-18 &\textbf{68.23}&8, 2&20.48& Zedboard \\ \hline
	\end{tabular}
	\footnotemark[1]{Xeon: Xeon E5-2630 v3;}
	\footnotemark[2]{1080Ti: Nvidia 1080Ti}
	\vspace{-9pt}
\end{table}

\subsubsection{Classification accuracy}
The classification accuracy on different datasets are shown in Table~\ref{tab:accuracy}.
For simplicity, we only show the top-1 accuracy.
Comparing to the floating point models (Floating in the table), the classification accuracy using ternary weights and quantized scalars and activations shows negligible degradation.
VecQ also achieves superior accuracy comparing to the recent works~\cite{efficient, li2016ternary}, in which only the weights are ternarized but not the scalars and activations.
Our proposed method shows better accuracy for Resnet-18 on ImageNet data set.
This result demonstrates the scalability and stability of VecQ, especially in aggressive low-bitwidth quantization scenarios.
%

\subsubsection{Model size reduction}
VecQ also greatly reduces the memory footprint (Mem. Reduc.) as shown in Table~\ref{tab:size}.
Ternary weight occupies only 2 bits whereas the original floating point requires 32 bits. 
As shown in Table~\ref{tab:size}, for convolution layers, VecQ compresses the parameters nearly to the theoretical limit (almost $16\times$ reduction). 
We quantize the last FC layer to 12-bit to maintain accuracy, so that the networks with less or no FC layers have higher compression ratio, such as Cifarnet and Resnet-18. Specifically, VecQ reduces up to 92.93\% (14.14$\times$) size of Resnet-18 in floating point.

\subsection{Ultra-low Latency Acceleration}

We use the models quantized by VecQ to evaluate our T-DLA accelerator design in terms of accuracy and frame per second (fps).
The measurements of the original models are on a server with two Intel Xeon E5-2630 v3 CPUs and one Nvidia 1080 Ti GPU.
T-DLA is implemented on a Xilinx Zedboard FPGA, which is suitable for edge applications with very limited logic resources. 
It has an on-chip dual-core ARM Cortex A9, and has 53.2K LUTs, 106.4K FFs, 140 BRAM blocks of 36Kb each, and 220 DSPs. Vivado System Design Suite 2019.2 is used for system implementation.

\subsubsection{Hardware Resource and Processing Latency Evaluation}
We choose an accelerator configuration that fully utilizes the given resources, shown in Table~\ref{tab:perandres},
together with the execution latency of the different models with this configuration.
We only show the most important configuration parameters including $T_{n}, T_{m}, L_{K}, L_{D}$, and the quantized bitwidth of the activations ($D_{w}$). 
As can be seen in Table~\ref{tab:perandres}, 
T-DLA with customized configuration can almost use all the resources, especially the DSPs.
The targeted FPGA can support up to 250MHz for the DSPs, which is twice of the frequency of other logic benefiting from the ternary computation array and independent clock design of the adder trees.

\subsubsection{Performance Comparison}
We compare T-DAL in terms of accuracy and fps with existing designs, either using the same DNN model or the same dataset. The results are shown in Table~\ref{tab:hocom}.
For MNIST dataset, the design in~\cite{finn} shows higher fps because of the DNN model they used is simpler and the ZC706 platform has almost $4\times$ more resources than ours. 
However, our implementation on Zedboard has a comparable fps (62051) to a design~\cite{impl16} (70000) with 3-bit weights on the ZC706 platform.
On CIFAR10 dataset, our design shows dominating accuracy advantage among all the VGG-like models.
On ImageNet dataset, 
we directly compare our results with the floating point version. T-DLA shows longer execution latency than the GPU but outperforms the CPU by $9.2\times$.

\section{Conclusions}
\label{sec:conclusion}
In this paper, we summarized our recent efforts for efficient on-device AI development including both training and inference.
We mainly focused on three major challenges of edge AI development.
First, we presented on-device training with ultra-low memory usage by proposing a novel rank-adaptive tensor-based tensorized
neural network model, which offers orders-of-magnitude memory reduction during training.
Second, we introduced VecQ, a novel quantization method that supports ultra-low bitwidth quantization with negligible accuracy degradation.
Third, we presented T-DLA, an ultra-low latency DNN accelerator design for ternarized DNNs achieving the state-of-the-art performance.
On top of the achievements in this paper, we expect more research breakthroughs to boost the development and deployment for the edge AI.

\bibliographystyle{ACM-Reference-Format}
\bibliography{ref}

\end{document}